\pdfoutput=1

\documentclass[11pt]{article}

\usepackage{acl}

\usepackage{times}
\usepackage{latexsym}

\usepackage[T1]{fontenc}

\usepackage[utf8]{inputenc}

\usepackage{microtype}

\usepackage{amsmath}
\usepackage{amssymb}
\usepackage{bm}

\usepackage{multirow}
\usepackage{graphicx}

\usepackage{float} 
\usepackage{subfigure}

\usepackage{amsthm}
\theoremstyle{definition}
\newtheorem{proposition}{Proposition}

\title{Improving Contextual Representation with Gloss Regularized Pre-training}

\author{Yu Lin$^*$,\ \ 
Zhecheng An$^*$,\ Peihao Wu,\ Zejun Ma \\
AI-Lab Speech \& Audio Team, ByteDance Inc., Beijing, China \\
\texttt{\{linyu.linyu,anzhecheng,wupeihao,mazejun\}@bytedance.com}}

\begin{document}
\maketitle

\def\thefootnote{*}\footnotetext{The first two authors contributed equally to this work.}
\def\thefootnote{\arabic{footnote}}

\begin{abstract}

Though achieving impressive results on many NLP tasks, 
the BERT-like masked language models (MLM) encounter the discrepancy between 
pre-training and inference. In light of this gap, we investigate the contextual representation 
of pre-training and inference from the perspective of word probability distribution. 
We discover that BERT risks neglecting the contextual word similarity in pre-training. 
To tackle this issue, we propose an auxiliary gloss regularizer module to BERT pre-training (GR-BERT), 
to enhance word semantic similarity. By predicting masked words and aligning contextual embeddings to 
corresponding glosses simultaneously, the word similarity can be explicitly modeled. We design two architectures 
for GR-BERT and evaluate our model in downstream tasks. Experimental results show that the gloss regularizer benefits 
BERT in word-level and sentence-level semantic representation. 
The GR-BERT achieves new state-of-the-art in lexical substitution task 
and greatly promotes BERT sentence representation 
in both unsupervised and supervised STS tasks.

\end{abstract}

\section{Introduction}

Pre-trained language models like BERT \cite{devlin-etal-2019-bert} and its variants 
\cite{Liu2019-roberta,lan2019albert,sun2019ernie,joshi2020spanbert} 
have achieved remarkable
success in a wide range of natural language processing (NLP) benchmarks.
By pre-training on large scale unlabeled corpora, BERT-like models learn contextual representations
with both syntactic and semantic properties. Researches show the contextual representations generated
by BERT capture various linguistic knowledge, including part-of-speech (PoS), 
named entities, semantic roles \cite{tenney2019probing,Liu2019,ettinger2020diagnostics}, 
word senses \cite{wiedemann2019wsd}, etc.  
Furthermore, with the fine-tuning procedure, the contextual representations show excellent transferability 
in downstream language understanding tasks, and lead to state-of-the-art (SOTA) performance.

The masked language model (MLM) plays a significant role in the pre-training stage of many BERT-like models 
\cite{Liu2019-roberta}.
In an MLM, a token $w$ is sampled from a text sequence $\textbf{s}$, and replaced with a \texttt{[MASK]} token. 
Let $\textbf{c}$ be the rest of tokens in $\textbf{s}$ except for $w$. 
We name $\textbf{c}$ as the \textit{masked context} or \textit{surrounding context}, 
and $\textbf{s}$ as the \textit{full context}.
During pre-training, BERT encodes the masked context $\textbf{c}$ into a contextual embedding vector $\bm{h}_c$,
and use it to generate a contextual token probability distribution $p(x|\textbf{c})$, where $x\in V$ and $V$
denotes the token vocabulary.  
The training objective is to predict the masked token $w$ by maximizing likelihood function $\log p(w|\textbf{c})$. 
In the fine-tuning or inference stage, BERT takes the 
full context $\textbf{s}$ without masks as input, 
and encodes every token into its contextual representation for downstream tasks.
We denote the contextual representation corresponds to token $w$ as $\bm{h}_s$. 

\begin{figure}[t] 
	\centering 
	\includegraphics[width=0.5\textwidth]{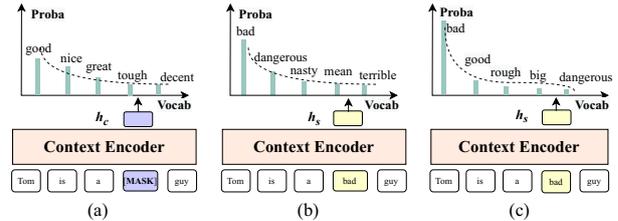}
	\caption{Conditional token probability distribution of tokens given masked context (a) 
	and full context (b) and (c). The ideal token distribution given full context is illustrated in (b), while (c)
	shows the full contextual token distribution generated by actual BERT.}
	\label{Fig_token_distribution} 
\end{figure}



We analyze the corresponding contextual token probability distribution
$p(x|\textbf{c})$ and $p(x|\textbf{s})$ generated from $\bm{h}_c$ and $\bm{h}_s$, as a proxy to
study the representations \cite{li2020sentence}. 
Figure \ref{Fig_token_distribution}(a) shows an example 
when masked context $\textbf{c}=$``\textit{Tom is a} \texttt{[MASK]} \textit{guy}'', 
the predicted tokens with high probabilities $p(x|\textbf{c})$ 
includes \textit{good}, \textit{nice}, \textit{great}, \textit{tough}, which are all reasonable 
answers to the Cloze task. 
Ideally, we want the context encoder to behave the same way when full context $\textbf{s}$ is given,  
as in Figure \ref{Fig_token_distribution}(b), the model should only propose
contextual synonyms of \textit{bad} such as \textit{dangerous}, \textit{nasty} and \textit{mean}
with $p(x|\textbf{s})$. However, the actual BERT generates $\hat{p}(x|\textbf{s})$ 
as shown in Figure \ref{Fig_token_distribution}(c), which contains inappropriate token proposals such
as \textit{good}, \textit{rough} and \textit{big}.

The discrepancy between Figure \ref{Fig_token_distribution}(b) and \ref{Fig_token_distribution}(c)
is because only the masked token distribution $p(x|\textbf{c})$ 
is explicitly modeled in BERT with the MLM, while the full contextual token distribution $p(x|\textbf{s})$ 
works in an agnostic way through model generalization. 
This leads to a gap between $p(x|\textbf{c})$
in pre-training and $p(x|\textbf{s})$ in fine-tuning and inference. 
It is shown in unsupervised semantic textual similarity (STS) tasks, 
BERT generates contextual embeddings that even underperforms
static embeddings for sentence representation \cite{reimers2019sentence}. 
Although in BERT pre-training, 
random token replacement strategy is used to mitigate the mismatch that \texttt{[MASK]} token is never seen
during fine-tuning, to the best of the authors' knowledge, 
there is no analysis on the gap of representation between masked context $\bm{h}_c$ and 
full context $\bm{h}_s$ in different phases when using BERT.  

To address this issue, we perform an investigation on 
the inner structure of $p(x|\textbf{s})$. 
Through theoretical derivation, we discover $p(x|\textbf{s})$
can be decomposed into the combination of masked contextual token distribution $p(x|\textbf{c})$ and 
a point-wise mutual information (PMI) term that describes contextual token similarity. 
Further analysis shows both the MLM and token replacement in BERT pre-training have potential
shortcomings in modeling the contextual token similarity. 
Inspired by the decomposition of $p(x|\textbf{s})$, we propose to add
an auxiliary gloss regularizer (GR) module to the MLM task, 
where mask prediction and gloss matching are trained simultaneously in the BERT pre-training.
We also design two model architectures to integrate the gloss regularizer into the original MLM task. 

We examine our proposed model in downstream tasks including unsupervised lexical substitution (LS) 
\cite{mccarthy-navigli-2007-semeval,kremer2014substitutes}, unsupervised STS 
and supervised STS Benchmark \cite{cer-etal-2017-semeval}. 
By invoking gloss regularized pre-training, our model improves lexical substitution task from $14.5$
to $15.2$ points in the LS14 dataset, 
leading to new SOTA performance.  
In unsupervised STS tasks, gloss regularizer improves the performance 
from $56.57$ to $67.47$ in terms of average Spearman correlation by a large margin. Such performance gain
is also observed in supervised STS task. 
Empirical experiments prove our model effectively generates better contextual token distribution and representations,
which contributes to word-level and sentence-level language understanding tasks.


\section{Related Works}
\paragraph{Masked Language Models.}
\citet{Liu2019-roberta} extend BERT into RoBERTa 
achieving substantial improvements. They claim the MLM task as the key contributor to 
contextual representation modeling, compared with 
next sentence prediction task. 
Many BERT variants focus on better masking strategies \cite{cui2019macbert,sun2019ernie,joshi2020spanbert}
to enhance the robustness
and transferability of contextual representative learning.
However, MLM suffers from the discrepancy between pre-training and fine-tuning since the $\texttt{[MASK]}$
tokens are only introduced during pre-training. 
To tackle this issue, permutation language model from XLNet \cite{Yang2019-xlnet}
and token replacement detection from ELECTRA \cite{clark2020electra} are proposed as 
alternative approaches to the MLM.
Instead of avoiding MLM, 
we analyze how the mask modeling affects the full contextual representation
in a probability perspective, and introduce gloss regularizer to mitigate the gap brought by MLM.

\paragraph{Contextual Representation Analysis.}
One way to analyze the contextual representation learned by pre-trained language model is through the probing tasks 
\cite{Liu2019,Miaschi2020,Vulic2020}, 
which are regarded as empirical proofs 
that pre-trained MLMs 
like BERT succeed in capturing linguistic knowledge.   
Many other researches focus on studying the geometry of contextual representations.
\citet{Ethayarajh2020} discovers anisotropy among the contextual embeddings of words
when studying contextuality of BERT. 
\citet{li2020sentence} propose a method using
normalizing flow to transform the contextual embedding
distribution of BERT into an isotropic distribution, and achieve performance gains in sentence-level tasks. 


\paragraph{Utilizing Word Senses.}
Because the BERT conveys contextualized semantic knowledge of polysemous, many researches use BERT
as a backbone to build word sense disambiguation (WSD) models 
\cite{huang2019glossbert,Blevins2020,bevilacqua2020breaking}. In these models, BERT is used as 
word senses and contexts encoders to perform the downstream matching task. 
One work that directly incorporates word sense knowledge into pre-training is 
SenseBERT \cite{levine2019sensebert} 
that introduces a weakly-supervised supersense prediction task, 
which leads to improvement on performance of WSD and word-in-context task. 
In SenseBERT, word prediction is enhanced with supersense category labels that act like
an external knowledge source. 
However, the gloss regularizer in our model provides fine-grained semantic information,
which aimed to align word representation space with the semantic space, and leads to better contextual
representations.

\section{Contextual Token Probability}
\label{sec:contexual_distribution}

\subsection{Masked Language Model}
Without loss of generality, the token probability distribution given full context $p(x|\textbf{s})$ 
can be decomposed into two parts,
\begin{align}
\log p(x|\textbf{s}) &= \log p(x|\textbf{c}) + \log\frac{p(x|w, \textbf{c})}{p(x|\textbf{c})} \nonumber \\
&=\log p(x|\textbf{c}) + \text{PMI}(x;w|\textbf{c}) \label{pmi_decomp}
\end{align}
where $\text{PMI}(x;w|\textbf{c})$ is the pointwise mutual information between $x$ and $w$ 
given $\textbf{c}$. 
PMI describes how frequently two tokens co-occur
than their independent occurrences, which is used as a measurement 
of the semantic similarity between tokens \cite{Ethayarajh2020,li2020sentence}.
In Eqn. (\ref{pmi_decomp}), $\log p(x|\textbf{c})$ only depends on masked context, 
which directly corresponds to the MLM training objective.
However, the PMI term is not explicitly modeled. 

In BERT, $p(x|\textbf{c})$ is generated from the encoded mask context $\bm{h}_c$ with a softmax
operation as
\begin{equation}\label{pxc_softmax}
p(x|\textbf{c})=\text{softmax}(\bm{h}_c^\top \bm{v}_x),
\end{equation}
where $\bm{v}_x$ stands for the embedding vector of token $x$ in vocabulary $V$. During fine-tuning or inference stage, 
full context $\textbf{s}$ without masks is encoded into $\bm{h}_s$ as the contextual representation of token $w$.
We can use the $\bm{h_s}$ to estimate $p(x|\textbf{s})$ 
in the same way as Eqn. (\ref{pxc_softmax}), denoted by $\hat{p}(x|\textbf{s})$,
\begin{equation}\label{pxs_softmax}
\hat{p}(x|\textbf{s})=\hat{p}(x|w,\textbf{c})=\text{softmax}(\bm{h}_s^\top \bm{v}_x).
\end{equation}

Under such approximation setup, $\text{PMI}(x;w|\textbf{c})$ can be transformed into
\begin{align}
\text{PMI}(x; w|\textbf{c})&\approx \log \frac{\hat{p}(x|w,\textbf{c})}{p(x|\textbf{c})}\nonumber\\
&=(\bm{h}_s-\bm{h}_c)^\top \bm{v}_x + \varphi(w,\textbf{c}), \label{pmi_mlm}
\end{align}
where $\varphi(w,\textbf{c})$ is constant w.r.t. $x$ (detailed
in Appendix \ref{appx:derivation}). In a deep neural network parameterized model like
BERT, $\bm{h}_s$ is encoded in an agnostic way. Thus, it is difficulty to further derive the PMI 
in Eqn. (\ref{pmi_mlm}). 

For a simpler case, if we consider a one-layer continuous bag-of-words (CBOW) model \cite{Mikolov2013}
\footnote{The CBOW model can be considered as a kind of masked language model.}, 
we have $\bm{h}_s-\bm{h}_c=\bm{h}_w$, where $\bm{h}_w$ is a context vector 
only related to the center token $w$. Now the PMI is formulated as
\begin{equation}\label{pmi_cbow}
\text{PMI}_{\text{CBOW}}(x; w|\textbf{c})=\log p(x|w) + \psi (w, \textbf{c}),
\end{equation}
where $\psi(w,\textbf{c})$ is another constant w.r.t. $x$ (also detailed
in Appendix \ref{appx:derivation}). In this case,
only the similarity information between $x$ and $w$ plays a role when comparing 
$\text{PMI}_{\text{CBOW}}(x; w|\textbf{c})$ among different candidate tokens $x\in V$, 
while the context information is irrelevant.

Although $\bm{h}_s-\bm{h}_c=\bm{h}_w$ is not satisfied in a deep model like BERT, the input sequences
for $\bm{h}_s$ and $\bm{h}_c$ share the most identical tokens $\textbf{c}$, 
and their only difference is whether to mask $w$.
Therefore, there is a potential risk that $\text{PMI}(x;w|\textbf{c})$ in MLM loses information related to 
the condition $\textbf{c}$, 
and degrades to the marginal $\text{PMI}(x;w)$, 
especially when the MLM lacks modeling $p(x|\textbf{s})$ in its training objective.

\subsection{Replaced Language Model}
In the BERT training process, a portion of tokens are replaced with random real tokens other than \texttt{[MASK]}, 
and the model is trained to predict the original tokens. 
We name this task as the replaced language 
model (RLM). Different from MLM, an RLM takes full context without 
masked tokens as input, and directly generates token 
distribution $p(x|\textbf{s})$, which seems to be a better way for full contextual representation modeling.  

We take a closer look at the RLM training process. 
Let $p(x|\textbf{s})=p(x|w,\textbf{c})$ be the probability that token $w$ is replaced with
token $x$ in context $\textbf{c}$. According to the Bayes' theorem, we have
\begin{equation}
p(x|w,\textbf{c})=\frac{p(x|\textbf{c})p(w|x,\textbf{c})}{\sum_{x'\in V} p(x'|\textbf{c}) p(w|x',\textbf{c})}.
\end{equation}

In a well-trained model, $p(w|x,\textbf{c})$ should be the replacing probability during training. 
Since the process of randomly replacing words is irrelevant to the context,
$p(w|x,\textbf{c})=p(w|x)$.
Let $\alpha$ be the probability when a token remains unchanged, and $1-\alpha$ be 
the replacing probability. Therefore,
\begin{equation}
  \label{rlm_prob}
  p(x|\textbf{s})=\frac{(1-\alpha)p(x|\textbf{c})}
  {\alpha |V| p(w|\textbf{c}) + (1-\alpha)\sum_{x'\neq w} p(x'|\textbf{c})},
\end{equation}
where $|V|$ denotes the vocabulary size. 
Eqn. (\ref{rlm_prob}) shows in RLM $p(x|\textbf{s})$ is proportional to $p(x|\textbf{c})$.
Therefore, $\text{PMI}(x;w|\textbf{c})$ is constant w.r.t. $x$, i.e.
\begin{align}
&\text{PMI}(x;w|\textbf{c})=\log\frac{p(x|w,\textbf{c})}{p(x|\textbf{c})}=\log\frac{p(x|\textbf{s})}{p(x|\textbf{c})} \nonumber\\
=&\log \frac{1-\alpha}{\alpha |V| p(w|\textbf{c}) + (1-\alpha)\sum_{x'\neq w} p(x'|\textbf{c})}. \label{rlm_pmi}
\end{align}

Combining Eqn. (\ref{rlm_pmi}) with Eqn. (\ref{pmi_decomp}), we conclude that 
the distribution of $x$ only relies on surrounding context $\textbf{c}$, 
but pays no attention to the center token $w$. This infers the RLM actually models the token distribution
conditioning on almost only the surrounding context, even if it takes full context as input.
As a result, the RLM fails to contribute better full contextual representation performance to the MLM,
since the PMI term, as a component part of the full contextual token distribution $p(x|\textbf{s})$, 
is completely ignored in the RLM. 
Solely using RLM would lead to worse contextual representation than using only MLM for pre-training.


\begin{figure*}[] 
	\centering 
	\subfigure[The framework of the gloss regularized BERT]{
		\includegraphics[width=0.8\textwidth]{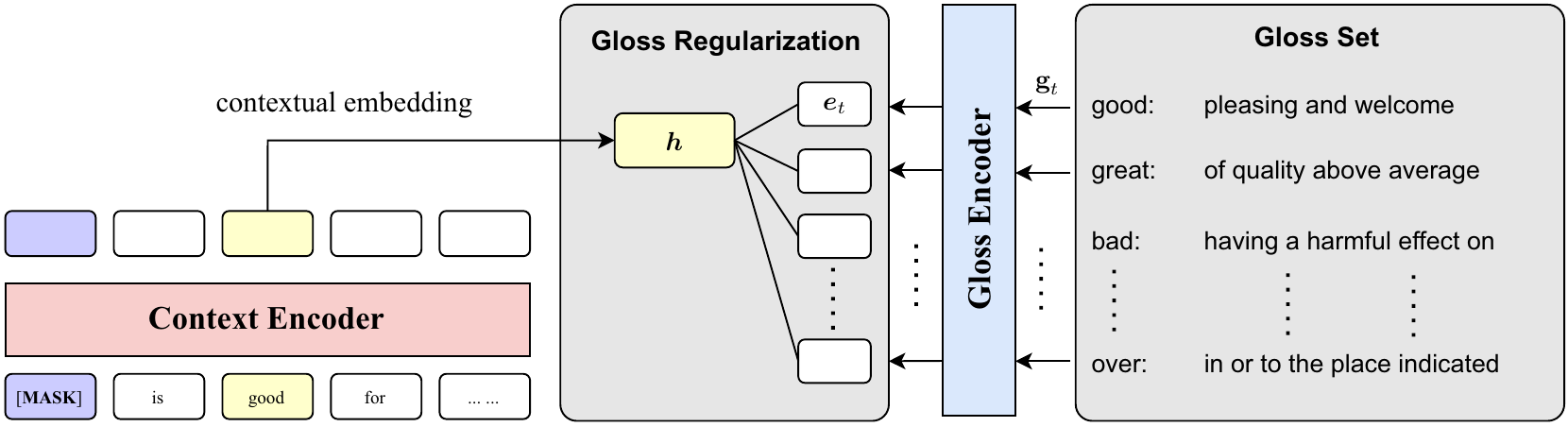}
	}
	\quad
	\subfigure[Two types of the context encoder structures]{
		\includegraphics[width=0.85\textwidth]{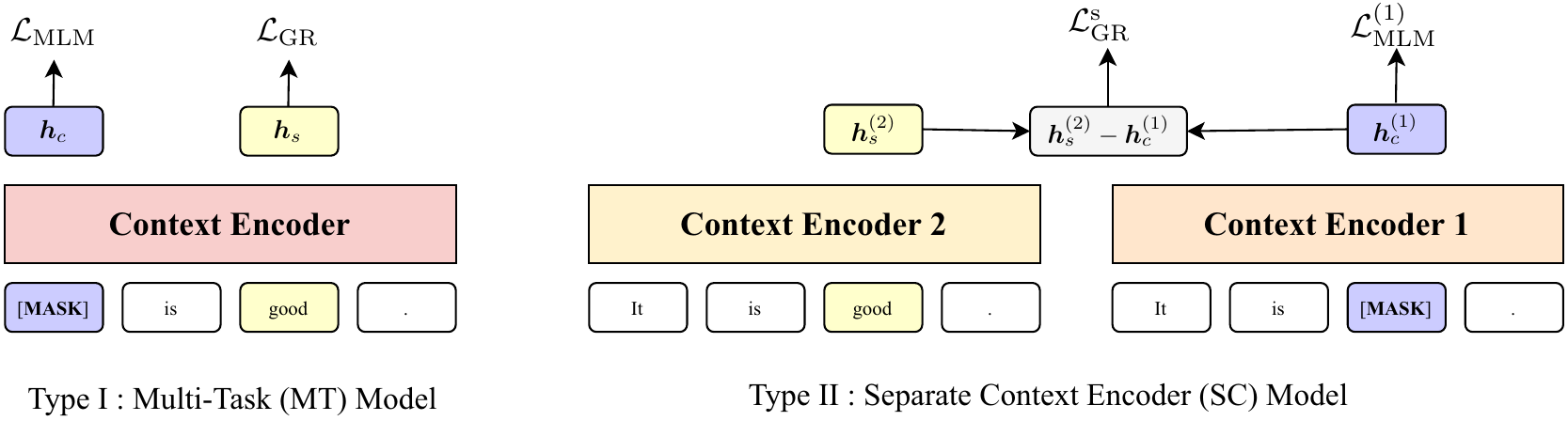}
	}
	\caption{ (a) shows the gloss regularizer aligns contextual representation space with the gloss space. 
	(b) Two GR architectures: the MT trains MLM and GR as multitask, while the SC
	utilizes two independent context encoders (the loss $\mathcal{L}_{\text{MLM}}^{(2)}$of SC is not shown). }
	\label{Fig_framework} 
\end{figure*}

\section{Gloss Regularizer}
\subsection{Invoking Gloss Matching}
As shown in Eqn. (\ref{pmi_decomp}), 
$p(x|\textbf{s})$ consists
of 
$p(x|\textbf{c})$ and $\text{PMI}(x;w|\textbf{c})$.  
Both MLM and RLM succeed in modeling $p(x|\textbf{c})$.
However, the analysis in Section \ref{sec:contexual_distribution} shows 
RLM completely ignores $\text{PMI}(x;w|\textbf{c})$, and MLM may suffer from potential risks that
the contextual information in $\text{PMI}(x;w|\textbf{c})$ would be lost, in either way the model
generates poor estimation of $p(x|\textbf{s})$.

$\text{PMI}(x;w|\textbf{c})$ describes co-occurrence probability of $x$ and $w$ normalized by their
marginal probabilities under context $\textbf{c}$ as condition. 
Ideally, it should be learned by training with labeled dataset 
$\{(\textbf{s}_1, \textbf{s}_2)\}$,
where $\textbf{s}_1=\{x_1, \textbf{c}\}$ and $\textbf{s}_2=\{x_2, \textbf{c}\}$ are semantically similar
text samples with shared context $\textbf{c}$ and exchangeable token pair $(x_1, x_2)$.  
However, such labeled data is expansive to build and not suitable for large-scale pre-training setup. 

Intuitively,
$\text{PMI}(x;w|\textbf{c})$ can be regarded as
semantic similarity between tokens under context. 
Although the contexts of similar tokens are hard to obtain, 
we can use the glosses of tokens as an alternative. 
Since the semantic of a word can be defined by its gloss, 
contextual token similarity can be determined by detecting whether tokens
are matching to similar glosses under context.
Therefore, 
in order to better
model the contextual token similarity defined by $\text{PMI}(x;w|\textbf{c})$,
we introduce gloss matching an auxiliary task named the \textit{gloss regularizer}. 
Two architectures to integrate gloss regularizer into MLM are detailed in Section \ref{sec:mt} and \ref{sec:sc}.

\subsection{Multitask Model}
\label{sec:mt}
A straight-forward method is to perform mask prediction and gloss matching as
joint multitasks (denoted as MT). In this architecture, the masked context $\textbf{c}$ and the 
full context $\textbf{s}$ are encoded by a context encoder into 
the contextual vector $\bm{h}_c$ and $\bm{h}_s$. 
The loss function of the MLM task is
\begin{equation}
\label{eq:mlm_loss}
\mathcal{L}_{\text{MLM}}=-\bm{h}_c^\top \bm{v}_w + \log \sum_{w'\in V}\exp(\bm{h}_c^\top \bm{v}_{w'}).
\end{equation}

For the gloss matching task, as illustrated in Figure \ref{Fig_framework}(a),
let $\textbf{g}_t$ be the gloss text of token $w$ under context $\textbf{c}$. 
Another gloss encoder is used to encode $\textbf{g}_t$ into a gloss vector $\bm{e}_t$.
Gloss matching is performed by calculating the similarity between the contextual token representation $\bm{h}_s$
and the gloss vector $\bm{e}_t$. The gloss regularizing loss is
\begin{equation}
\label{eq:gr_loss}
\mathcal{L}_{\text{GR}}=-\text{sim}[\bm{h}_s, \bm{e}_t] + \log \sum_{t'\in T}\exp\text{sim}[\bm{h}_s, \bm{e}_{t'}],
\end{equation}
where $\text{sim}[\cdot,\cdot]$ is a similarity measurement function, and $T$ is a set of negative glosses. 
The final loss function is the combination of the two losses, 
\begin{equation}
\mathcal{L}_{\text{MT}}=\mathcal{L}_{\text{MLM}}+\lambda \mathcal{L}_{\text{GR}},
\end{equation}
where $\lambda$ denotes the regularizing weight.

This setting resembles the bi-encoders model (BEM) for WSD proposed by \cite{Blevins2020}. 
However, in our model, the context encoder is trained on mask prediction
task simultaneously with the gloss matching task, while the BEM takes gloss matching as a fine-tuning task.
We train the two tasks together 
for better contextual and semantic representation modeling. 
As a result, the model learns token distribution not only conditioning on the masked context,
but also influenced by semantic similarity with center token, 
which gives a better estimation of $p(x|\textbf{s})$.

\subsection{Separate Context Encoder Model}
\label{sec:sc}
Another method is directly inspired by the decomposition from Eqn. (\ref{pmi_decomp}). 
Different from the multitask model, we use two context encoders instead of one (denoted as SC). 
The first context encoder, denoted by $\texttt{enc}_1$, encodes the masked context as
$\bm{h}_c^{(1)}=\texttt{enc}_1(\textbf{c})$, and 
learns purely from MLM task with loss $\mathcal{L}_{\text{MLM}}^{(1)}$ derived similar as
Eqn. (\ref{eq:mlm_loss}).

The full context $\textbf{s}$ is encoded into $\bm{h}_s^{(2)}=\texttt{enc}_2(\textbf{s})$
by the second context encoder. 
Eqn. (\ref{pmi_mlm}) shows $\text{PMI}(x; w|\textbf{c})$ is entailed
in the linear difference between the encoding of full and masked context. 
Therefore, we use
$(\bm{h}_s^{(2)}-\bm{h}_c^{(1)})$ for gloss matching, where the loss function
is formulated as
\begin{multline}
\mathcal{L}_{\text{GR}}^{\text{s}}=-\text{sim}\big[\bm{e}_t, \bm{h}_s^{(2)}-\bm{h}_c^{(1)}\big] \\
+\log \sum_{t'\in T}\exp\text{sim}\big[\bm{e}_{t'}, \bm{h}_s^{(2)}-\bm{h}_c^{(1)}\big].
\end{multline}
In order to make the gloss matching learned by $\texttt{enc}_2$ aligned with the word embedding space,
another MLM task is added to the training of $\texttt{enc}_2$, 
with loss $\mathcal{L}_{\text{MLM}}^{(2)}$. 
Thus, the complete loss function of the SC model is
\begin{equation}
\mathcal{L}_{\text{SC}}=\mathcal{L}_{\text{MLM}}^{(1)}+
\mathcal{L}_{\text{MLM}}^{(2)}+\lambda \mathcal{L}_{\text{GR}}^{\text{s}}.
\end{equation}

Although one gloss encoder and two contextual encoders are involved during training, 
only $\texttt{enc}_2$ is used at the inference stage. 
The estimation of contextual token distribution is 
given by $\hat{p}(x|\textbf{s})=\text{softmax}(\bm{v}_w^\top \bm{h}_s^{(2)})$. By using two separate
contextual encoders, the MLM task and the gloss matching task can be trained individually, 
which leads to better performance for each task. 
Besides, the combination of the two tasks corresponds to the theoretical derivation of $p(x|\textbf{s})$,
and integrates the gloss regularizer in a more natural and explainable way.

\subsection{Gloss Regularized Pre-training}



To pre-train the GR-BERT, we employ the gloss dataset from the online Oxford dictionary released by \citet{chang2018xsense,chang-chen-2019-word}, 
formatted in triplets: word, sentence and definition ($word_i$, $sent_{ij}$, $def_{ik}$), where the human understandable gloss text $def_{ik}$ describes the sense of target $word_i$ in the sentence context $sent_{ij}$. 
The data consists of 677,191 pieces in total, including 31,889 words and 78,105 glosses.   

We train the GR-BERT model with the gloss matching loss. 
For each triplet sample, context $sent_{ij}$ with $word_i$ is encoded by the context encoder, 
and the target gloss $def_{ik}$ is encoded by the gloss encoder. 
The cosine function is used as the similarity function $\text{sim}[\cdot,\cdot]$ 
in Eqn. \eqref{eq:gr_loss}.
To build the negative gloss set in training, we use the in-batch negative sampling strategy \cite{chen2017sampling}. For each triplet sample ($word_i$, $sent_{ij}$, $def_{ik}$) in a batch, the positive glosses of other samples in the batch make up the negative gloss set $T$ for the target word $word_i$. 
Since it would be relatively easy to distinguish the positive gloss $def_{ik}$ from the randomly build-up in-batch negative set, we add the hard negative gloss to the negative gloss set. For each triplet ($word_i$, $sent_{ij}$, $def_{ik}$), we randomly pick another gloss $def_{il}$ of the target word $word_i$ as the hard negative.  

We utilize the BERT and RoBERTa models to initialize the context and gloss encoders, both of which are pre-trained on the gloss dataset for about 10 epochs. The gloss-matching accuracy is used as the metric to evaluate the model performance in pre-training. 
Detailed pre-training settings and hyper-parameters are provided in Appendix \ref{appx:training_details}.


\section{Experiments}

\subsection{Downstream Tasks} 
In this section, we evaluate our model on three language understanding tasks. 
First, we choose the lexical substitution task to observe the word-level semantic performance. 
Then we conduct experiments on two sentence representation tasks: 
the STS task in unsupervised setting and the supervised STS benchmark (STS-B) task.


\subsection{Lexical Substitution} 

\begin{table*}[]
	\centering
	\resizebox{\textwidth}{!}{%
		\begin{tabular}{l|l|l|ll|ll|l|l}
			\hline
			\multirow{2}{*}{Method} & \multirow{2}{*}{Backbone} & \multirow{2}{*}{Post-Process} & \multicolumn{3}{c|}{SemEval 2007 (LS07)} & \multicolumn{3}{c}{CoInCo (LS14)} \\ \cline{4-9} 
			&  &  & \multicolumn{1}{c|}{\textbf{best}/best-m} & \multicolumn{1}{c|}{oot/oot-m} & \multicolumn{1}{c|}{P@1/P@3} & \multicolumn{1}{c|}{\textbf{best}/best-m}  &   \multicolumn{1}{c|}{oot/oot-m} & \multicolumn{1}{c}{P@1/P@3} \\ \hline

			\citet{roller-erk-2016-pic} & SGNS emb& - & \multicolumn{1}{l|}{-} &  \multicolumn{1}{l|}{-}  & \multicolumn{1}{l|}{19.7/14.8} & \multicolumn{1}{l|}{-} &  \multicolumn{1}{l|}{-}  & 18.2/13.8  \\ 
			
			\citet{zhou-etal-2019-bert}   & BERT$_{\text{large}}$ & - & \multicolumn{1}{l|}{12.1/20.2} &  \multicolumn{1}{l|}{40.8/56.9} &  \multicolumn{1}{l|}{13.1/-} & \multicolumn{1}{l|}{9.1/19.7} &  \multicolumn{1}{l|}{33.5/56.9} & 14.3/- \\ 
			
			&  & +valid & \multicolumn{1}{l|}{20.3/34.2} &  \multicolumn{1}{l|}{55.4/68.4} & \multicolumn{1}{l|}{51.1/-} & \multicolumn{1}{l|}{14.5/33.9} &  \multicolumn{1}{l|}{\textbf{45.9}/69.9} & \textbf{56.3}/- \\ 
			
			\citet{arefyev-etal-2020-always} & RoBERTa$_{\text{large}}$ & - & \multicolumn{1}{l|}{-} &  \multicolumn{1}{l|}{-} & \multicolumn{1}{l|}{32.0/24.3} & \multicolumn{1}{l|}{-} &  \multicolumn{1}{l|}{-} & 34.8/27.2 \\ 
			
			&  & +emb & \multicolumn{1}{l|}{-} &  \multicolumn{1}{l|}{-} & \multicolumn{1}{l|}{44.1/31.7} & \multicolumn{1}{l|}{-} &  \multicolumn{1}{l|}{-} & 46.5/36.3 \\ 
			& XLNet$_{\text{large}}$ & +emb & \multicolumn{1}{l|}{-} &  \multicolumn{1}{l|}{-} & \multicolumn{1}{l|}{49.5/34.9} & \multicolumn{1}{l|}{-} &  \multicolumn{1}{l|}{-} & 51.4/39.1 \\ 
			\hline
			
			Baselines & BERT$_{\text{base}}$ & - & \multicolumn{1}{l|}{13.2/22.3} & \multicolumn{1}{l|}{40.8/57.1} & \multicolumn{1}{l|}{33.1/23.7} & \multicolumn{1}{l|}{10.1/21.9} & \multicolumn{1}{l|}{33.0/56.5} & 38.4/28.7 \\ 	
			
			
			& RoBERTa$_{\text{base}}$ & - & \multicolumn{1}{l|}{16.7/27.8} &  \multicolumn{1}{l|}{45.2/62.9} & \multicolumn{1}{l|}{40.8/28.5}  & \multicolumn{1}{l|}{11.0/23.6} & \multicolumn{1}{l|}{34.9/59.3} & 42.2/31.4  \\ 

			
			\hline
			
			Our work & MT GR-BERT$_{\text{base}}$  & - & \multicolumn{1}{l|}{17.7/30.8} &  \multicolumn{1}{l|}{49.8/67.8} & \multicolumn{1}{l|}{42.5/31.1 }& \multicolumn{1}{l|}{12.2/ 26.5} & \multicolumn{1}{l|}{39.2/64.5} & 46.4/35.3 \\ 
			& SC GR-BERT$_{\text{base}}$ & - & \multicolumn{1}{l|}{18.2/31.2} &  \multicolumn{1}{l|}{49.9/67.6} & \multicolumn{1}{l|}{44.1/31.2} & \multicolumn{1}{l|}{12.4/ 27.1} &  \multicolumn{1}{l|}{39.8/65.5} &46.6/35.8 \\
			
			& MT GR-RoBERTa$_{\text{base}}$ & - & \multicolumn{1}{l|}{19.7/32.9} & \multicolumn{1}{l|}{53.0/72.8} & \multicolumn{1}{l|}{47.9/34.2} & \multicolumn{1}{l|}{12.9/28.3} & \multicolumn{1}{l|}{40.6/66.4} &48.6/37.2 \\ 
			
			& SC GR-RoBERTa$_{\text{base}}$  & - & \multicolumn{1}{l|}{19.4/33.2} &  \multicolumn{1}{l|}{52.8/71.5} & \multicolumn{1}{l|}{47.4/33.4} & \multicolumn{1}{l|}{13.1/28.8} & \multicolumn{1}{l|}{40.9/66.6} & 48.8/37.8 \\ 
				
			&  & +emb & \multicolumn{1}{l|}{22.4/38.2} &  \multicolumn{1}{l|}{56.4/76.0} & \multicolumn{1}{l|}{53.7/37.8} & \multicolumn{1}{l|}{14.5/32.8} & \multicolumn{1}{l|}{43.8/69.9} & 53.5/ 41.4 \\ 
			
			&  & +valid & \multicolumn{1}{l|}{22.6/38.4} &  \multicolumn{1}{l|}{56.0/73.9} & \multicolumn{1}{l|}{54.8/39.0} & \multicolumn{1}{l|}{15.1/33.7} & \multicolumn{1}{l|}{44.1/69.6} & 56.0/42.7 \\ 
			
			&  & +both & \multicolumn{1}{l|}{\textbf{23.1/39.7}} & \multicolumn{1}{l|}{\textbf{57.6/76.3}} & \multicolumn{1}{l|}{\textbf{55.0/40.3}} & \multicolumn{1}{l|}{\textbf{15.2/34.4}} & \multicolumn{1}{l|}{45.3/\textbf{71.3}} &  55.9/\textbf{43.5} \\ \hline
			
		\end{tabular}%
	}
	\caption{Comparison with previous SOTA on lexical substitution task. 
	Results of the first three works are from the mentioned papers and the results 
	in the baseline are from our experiments with the same word process. }
	\label{tab:LS-result}
\end{table*}

\paragraph{Task and Dataset.} 
Lexical substitution aims to replace the target word in a given context sentence 
by a substitute word that not only is semantically consistent with the original word 
but also preserves the sentence's meaning. 
There are two benchmark datasets for this task: the SemEval 2007 dataset (LS07) \cite{mccarthy-navigli-2007-semeval} 
with 201 target words, and the CoInCo dataset (LS14) \cite{kremer2014substitutes} with 4,255 target words, 
both of which are unsupervised.
The task LS07 releases the official evaluation metrics \textit{best/best-mode} 
and \textit{oot/oot-mode}\footnote{\url{http://www.dianamccarthy.co.uk/task10index.html}}, 
which evaluate the quality of the best prediction and the best 10 predictions, separately. 
We also report the metrics precision@1 (P@1) and P@3. 
Because the metric \textit{best} considers the word frequencies in annotated labels, 
we take it as the main metric in this task.

\paragraph{Candidate Generation.} We use the context encoder pre-trained with GR to generate lexical substitutions. 
Given a target word $w$ and its context $\textbf{s}$, 
we directly employ the full contextual token distribution $p(x|\textbf{s})$ to perform the word prediction, then sort the candidates by their probabilities. 

Before evaluating the score of the generated candidates, we filter out the words with the same lemmatization or with different PoS from the target word. All our experiments employ the same candidate filtering process. The detailed process is in Appendix \ref{appx:ls_example}.

\paragraph{Post-Process.} 
Previous works proposed several effective approaches to improve LS performance. 
\citet{arefyev-etal-2020-always} used the input word embedding 
to inject more target word information (noted \textit{+emb}). 
\citet{zhou-etal-2019-bert} utilized a pre-trained model to re-score candidates (noted \textit{+valid}). 
We denote these approaches as \textit{post-process} and adopt them in our experiments. 
As \citet{arefyev-etal-2020-always} reported, 
the result in \cite{zhou-etal-2019-bert} is hardly reproduced and their code is not available, 
we then implement the validation process by ourselves. 

\paragraph{Result and Analysis.}
Table~\ref{tab:LS-result} shows the comparison of our models with the previous SOTAs in LS07 and LS14 benchmarks. We use publicly released BERT and RoBERTa models as the baseline to generate lexical candidates as illustrated in the candidate generation process. Initialized by the same BERT or RoBERTa parameters, our GR-models are further pre-trained on gloss dataset with GR module.
We first compare the model outputs without post-process. 
Our GR models surpass their MLM baselines by large margins in all metrics: 
the \textit{best} value increases more than 3 points, the \textit{oot} increases about 8 points in LS07. 
In separate context encoder structure, 
the \textit{best} value of BERT increases from $10.1$ to $12.4$ in LS14, 
and the metric increases from $11.0$ to $13.1$ for RoBERTa. 
Comparing the P@1 with \cite{arefyev-etal-2020-always}, 
the SC GR-RoBERTa base model $48.8$ even exceeds the large RoBERTa model with \textit{emb} $46.5$. 

Results indicate that GR model generates more semantically similar words 
and preserve the sentence original meaning even though no LS-like training data is used. 
This is because the gloss regularization plays the key role in modeling contextual token
distribution $p(x|\textbf{s})$ by taking both contextual and semantic information into consideration. 
Given a sentence context, if two words are semantically replaceable, 
their gloss text descriptions are naturally similar. As the word contextual embedding is aligned with its gloss, 
the words in semantically similar contexts are gathered closer indirectly, which benefits the LS task.

We further apply post-process on the SC GR-RoBERTa model. 
Consistent with previous works \cite{arefyev-etal-2020-always, zhou-etal-2019-bert}, 
both processes improve the performance in testset LS14: $+emb$ increases the \textit{best} value from 13.1 to 14.5, 
and it is to 15.1 using $+valid$. By applying both post-processes, 
our SC GR-RoBERTa model achieves the new SOTA 15.2 in \textit{best}. 
We also achieve SOTA in the metrics \textit{best-m/oot-m} and P@3 in LS14 and all metrics in LS07. 
Appendix \ref{appx:ls_example} demonstrates random selected examples of the LS task and the model outputs.

\subsection{Unsupervised Sentence Representation Task}

\begin{table*}[t]
	\centering
	\resizebox{0.95\textwidth}{!}{%
		\begin{tabular}{lllllllll}
			\hline
			Model & STS12 & STS13 & STS14 & STS15 & STS16 & STS-B & SICK-R & Avg. \\ 
			\hline
			
			GloVe embs & 55.14 & 70.66 & 59.73 & 68.25 & 63.66 & 58.02 & 53.76 & 61.32 \\
			BERT-flow & 58.40 & 67.10 & 60.85 & 75.16 & 71.22 & 68.66 & 64.47 & 66.55 \\
			BERT-whitening(NLI) & 57.83 & 66.90 & 60.90 & 75.08 & 71.31 & 68.24 & 63.73 & 66.28 \\
			SimCSE-BERT & 68.40 & \textbf{82.41} & \textbf{74.38} & 80.91 & 78.56 & 76.85 & \textbf{72.23} & 76.25 \\ 
			SimCSE-RoBERTa & \textbf{70.16} & 81.77 & 73.24 & \textbf{81.36} & \textbf{80.65} & \textbf{80.22} & 68.56 & \textbf{76.57} \\
			\hline
			
			BERT(first-last avg) & 39.70 & 59.38 & 49.67 & 66.03 & 66.19 & 53.87 & 62.06 & 56.70 \\
			MT GR-BERT(first-last avg.) & 53.20 & \textbf{69.68} & 58.81 & \textbf{73.25} &  \textbf{72.16} & \textbf{66.65} & \textbf{66.47} & \textbf{65.75} \\
			SC GR-BERT(first-last avg.) & \textbf{53.69} & 68.66 & \textbf{58.83} & 71.90 & 71.64 & 66.18 & 66.46 & 65.34 \\ 
			\hline
			
			RoBERTa(first-last avg.) & 40.88 & 58.74 & 49.07 & 65.63 & 61.48 & 58.55 & 61.63 & 56.57 \\
			MT GR-RoBERTa(first-last avg.) & \textbf{53.73} & \textbf{72.57} & \textbf{61.04} & \textbf{75.23} & \textbf{72.86} & 69.44 & \textbf{67.39} & \textbf{67.47} \\
			SC GR-RoBERTa(first-last avg.) & 53.69 & 70.00 & 59.24 & 72.38 & 72.47 & \textbf{70.12} & 67.02 & 66.42 \\ 
			\hline
		\end{tabular}%
	}
\caption{Sentence embedding performance on unsupervised STS tasks. 
Results in the first row are from \citealt{gao2021simcse}. 
Notation (first-last avg) means take the average of word embs from the input and output layer.
}
\label{tab:sts-result}
\end{table*}

\paragraph{STS Task and Dataset.} 
STS tasks deal with determining how similar two sentences are. 
We evaluate our model on 7 STS tasks: STS tasks 2012-2016 
\cite{agirre2012semeval, agirre2013sem, agirre-etal-2014-semeval, agirre-etal-2015-semeval,agirre2016semeval}, 
STS Benchmark (STS-B) \cite{cer-etal-2017-semeval} and SICK-Relatedness (SICK-R) \cite{marelli2014sick}. 
Following the work of \citet{gao2021simcse} and their setting 
in STS tasks\footnote{\url{https://github.com/princeton-nlp/SimCSE}},
we use \textit{Spearman's correlation} with \textit{``all'' aggregation} as the evaluation metric,
and use no additional regressor in experiments.


\paragraph{Baselines.} 
Since our experiments are unsupervised w.r.t. STS task: neither STS data 
nor NLI dataset\footnote{NLI dataset consists of SNLI and MNLI, 
both of which are proved to be effective domain data for STS tasks \cite{gao2021simcse, reimers2019sentence}.}
are used for training, we only perform comparison with previous works in unsupervised setting. 
SOTA works for these tasks are either trained by carefully designed sentence-level loss 
[e.g. SimCSE \cite {gao2021simcse}, BERT-flow \cite{li2002learning}] or tuned on sentence dataset NLI 
[e.g. BERT-whitening \cite{su2021whitening}]. 
Therefore, these models are able to generate effective sentence representation. 
In contrast, our model is not trained with any sentence tasks, 
and we simply use the average of contextual word embeddings to represent sentence. 
Thus, it is not very fair to directly compare with the mentioned sentence encoders. 
We then focus more on the comparison with the original MLM.

\paragraph{Result and Analysis.} 
Table~\ref{tab:sts-result} shows the results on STS tasks. We employ the publicly released BERT and RoBERTa pre-trained models as the baselines, and our GR-models are further pre-trained with GR-module in the gloss dataset. 
With gloss regularization in pre-training, the average Spearman's correlation increases from $56.70$ to $65.75$ 
in BERT model and from $56.57$ to $67.47$ for RoBERTa. 
Though still far below the SimCSE SOTA performance, 
our model approaches the BERT-whitening and BERT-flow 
without any deliberately designed sentence-level tasks or transforming word distribution on domain data. 
\citet{reimers2019sentence} report the unsupervised BERT embedding 
is infeasible for STS and performs even worse than GloVe embedding. 
\citet{li2020sentence} blame it on the anisotropic distribution of BERT word embeddings. 
Our experiments show great gains of GR-BERT in sentence embedding, 
proving the advantage of gloss regularized contextual representation is also valid for sentences. 
A brief analysis on sentence representation with gloss regularizer is provided in Appendix \ref{appx:sent_similarity}.

\subsection{Supervised STS}
\paragraph{STS-B Task and Dataset.} 
We validate our model in supervised STS Benchmark (STS-B) \cite{cer-etal-2017-semeval}. 
The data consists of 8,628 sentence pairs and is divided into trainset (5,749), devset (1,500) and testset (1,379). 

Since supervised STS performance are largely influenced by the training data, 
we only use the STS trainset in all experiments. 
Besides, we randomly reduce the data size to simulate the limit data scenarios 
and compare our model with MLM baselines. 
Following the sentence-BERT 
\cite{reimers2019sentence}\footnote{\url{https://www.sbert.net/examples/training/sts/README.html}}, 
we use Siamese BERT network with cosine similarity.

\paragraph{Result and Analysis.} 

\begin{table}[]
	\centering
	\resizebox{0.4\textwidth}{!}{%
		\begin{tabular}{clc}
			\hline
			Data ratio & Models & Spearman \\ \hline
			$100\%$ & BERT &  $ 83.98 \pm 0.16$ \\
			& MT GR-BERT& $ \textbf{85.13} \pm 0.06 $ \\
			& SC GR-BERT & $ 85.00 \pm 0.16$ \\ 
			\hline
			$100\%$ & RoBERTa & $85.90 \pm 0.57$ \\
			& MT GR-RoBERTa & $\textbf{86.87} \pm 0.21$ \\
			& SC GR-RoBERTa & $86.25 \pm 0.30$ \\
			\hline
			$50\%$ & BERT & $ 81.60 \pm 0.28 $ \\
			& MT GR-BERT & $ \textbf{83.47} \pm 0.15 $ \\
			& SC GR-BERT& $ 83.06 \pm 0.19 $ \\
			\hline
			$20\%$ & BERT & $ 76.43 \pm 0.37 $  \\
			& MT GR-BERT & $ \textbf{79.87} \pm 0.41 $ \\
			& SC GR-BERT& $ 79.18 \pm 0.21 $ \\
			\hline
		\end{tabular}%
	}%
 \caption{Evaluation on STS-B test set. All experiments are fine-tune for 4 epochs with batch size 16. 
 Results are the average of 4 random seeds.}
 \label{tab:stsb}
\end{table}


Tabel~\ref{tab:stsb} shows the comparison on STS-B. In both BERT and RoBERTa backbones, 
GR models improve the baselines by around $0.9$ points. In low-resource scenarios, 
the advantage of GR-BERT increases. When $50\%$ data is available, 
the gain of MT GR-BERT is increased to 1.87 points, and the gain is up to 3.44 points for $20\%$ data. 
Results show that in fine-tuning process, the GR model still preserves its advantage over MLM baselines 
in sentence semantic representation, indicating the contextual representation pre-trained with GR is 
transferable in further fine-tuning.
The GR pre-training is able to enhance the semantic knowledge in model, 
especially in the low-resource data scenarios,  
which ease the hunger for task training data.

\begin{table}[]
	\centering
	\resizebox{0.4\textwidth}{!}{%
		\begin{tabular}{lccc}
			\hline
			model & \multicolumn{1}{c}{LS14} & \multicolumn{1}{c}{STS Avg} & \multicolumn{1}{c}{STS-B} \\ \hline
			BERT & 10.1 & 56.70 &  $ 83.98  $ \\ 	
			+MLM & 10.9 & 62.22 & $ 84.62$ \\	
		MT GR-BERT & 12.2 & 65.75 &  $ 85.13$   \\ 		
		SC GR-BERT & 12.4 & 65.34 &  $ 85.00$   \\ \hline	
		\end{tabular}%
	}%
 \caption{Ablation studies of different training loss in three tasks. +MLM means only use MLM loss in training. 
 We use the metric  \textit{best} for LS14 task, the average Spearman's correlation for 7 STS tasks and STS-B.} 
  \label{tab:ablation}
\end{table}

\subsection{Ablation Analysis}
We now investigate the influence of gloss training data and the model structures. 
Results are shown in Table~\ref{tab:ablation}. 
\citet{gururangan2020don} reports the domain data pre-training can improve model performance. 
To evaluate the influence of dictionary corpus, 
we pre-train BERT by MLM in the same dataset and find that high-quality data improves all three task performances. 
However, GR still contributes to the large part of the improvement, especially in the LS task. 
As for the two proposed structures, 
the SC-GR utilizes individual context encoders  
that impose less restriction on gloss learning, and achieves better performance in LS word-level task. 
On the contrary, 
the MT model provides a better sentence embedding and surpasses SC structure in STS tasks.


\section{Conclusion}

In this work, we propose the GR-BERT, a model with gloss regularization to enhance the word contextual information. 
We first analyze the gap between MLM pre-training and inference, 
and aim to model the PMI term that characterizes the word semantic similarity given context. 
Due to the lack of data that labels the word semantic similarities given contexts, 
we propose to indirectly learn the semantic information in pre-training 
by aligning contextual word embedding space to a human annotated gloss space. 
We design two model structures and validate them in three NLP semantic tasks. 
In the lexical substitution task, we increase the SOTA value from $14.5$ to $15.2$ in LS14 \textit{best} metric 
and many other metrics in LS07 and LS14 are also improved. In the unsupervised STS task, 
our GR model show its capacity in sentence representation without any training in sentence task, 
and it improves the MLM performance from $56.57$ to $67.47$. 
In the supervised STS-B task, GR model exceed the MLM baseline by about 0.9 points, 
and the gains increases to 3.44 in the low resource scenarios.

Our work provides a new perspective to the MLM pre-training, 
and show the effectiveness of modeling word semantic similarity. 
However, one limitation of our work is the lack of large-scale word-gloss matching data. The training data in our work is far less than that in BERT pre-training, which hinders the large-scale pre-training in GR-BERT. Our future works will focus on mining more word-gloss training data and validate GR model in more NLP tasks. We believe there is still a big room for GR model performance improvement and possible gains in more NLP tasks.


\bibliography{anthology,custom}
\bibliographystyle{acl_natbib}
\newpage

\appendix
\section{Derivation of Eqn. (\ref{pmi_mlm}) and (\ref{pmi_cbow})}
\label{appx:derivation}
By plugging Eqn. (\ref{pxc_softmax}) and (\ref{pxs_softmax}) into Eqn. (\ref{pmi_mlm}), we have
\begin{align}
\text{PMI}(x; &w|\textbf{c})\approx\log\frac{\hat{p}(x|w,\textbf{c})}{p(x|\textbf{c})}\nonumber\\
=&\log\frac{\text{softmax}(\bm{h}_s^\top \bm{v}_x)}
{\text{softmax}(\bm{h}_c^\top \bm{v}_x)} \nonumber\\
=&\log \frac{e^{\bm{h}_s ^\top \bm{v}_x}}{\sum_{x'} e^{\bm{h}_s^\top \bm{v}_{x'}}} - 
\log \frac{e^{\bm{h}_c^\top \bm{v}_x}}{\sum_{x'} e^{\bm{h}_c^\top \bm{v}_{x'}}}\nonumber\\
=& (\bm{h}_s-\bm{h}_c)^\top \bm{v}_x  +
\log \frac{\sum_{x'}  e^{\bm{h}_c^\top \bm{v}_{x'}} }{\sum_{x'} e^{\bm{h}_s^\top \bm{v}_{x'}}}. \label{pmi_detail1}
\end{align}
The second term in Eqn. (\ref{pmi_detail1}) can be denoted as
\begin{equation*}
\varphi(w, \textbf{c})=
\log \frac{\sum_{x'}  e^{\bm{h}_c^\top \bm{v}_{x'}} }{\sum_{x'} e^{\bm{h}_s^\top \bm{v}_{x'}}},
\end{equation*}
since $\varphi(w, \textbf{c})$ is a function w.r.t. only $w$ and $\textbf{c}$, and is constant to $x$.

For the CBOW model, by applying $\bm{h}_w=\bm{h}_s-\bm{h}_c$, the PMI function given by Eqn. (\ref{pmi_detail1}) 
can be transformed into
\begin{align}
&\text{PMI}_{\text{CBOW}}(x; w|\textbf{c}) 
\approx\bm{h}_w^\top \bm{v_x} + 
\log \frac{\sum_{x'}  e^{\bm{h}_c^\top \bm{v}_{x'}} }{\sum_{x'} e^{\bm{h}_s^\top \bm{v}_{x'}}}\nonumber\\
=&\log \frac{e^ {\bm{h}_w ^\top \bm{v}_x}}{\sum_{x'} e^{\bm{h}_w^\top \bm{v}_{x'}}}
+\log \frac{\sum_{x'}  e^{\bm{h}_c^\top \bm{v}_{x'}} \sum_{x'} e^{\bm{h}_w^\top \bm{v}_{x'}}}
{\sum_{x'} e^{\bm{h}_s^\top \bm{v}_{x'}}} \nonumber\\
=&\log p(x|w) + \psi(w,\textbf{c}),
\end{align}
where
\begin{equation*}
\psi(w,\textbf{c})=
\log \frac{\sum_{x'}  e^{\bm{h}_c^\top \bm{v}_{x'}} \sum_{x'} e^{\bm{h}_w^\top \bm{v}_{x'}}}
{\sum_{x'} e^{\bm{h}_s^\top \bm{v}_{x'}}}.
\end{equation*}
$\psi(w,\textbf{c})$, like $\varphi(w,\textbf{c})$, is also constant w.r.t. $x$.

\section{Pre-training Details}
\label{appx:training_details}

We employ the BERT-base uncased model and RoBERTa-base model to initialize the context and gloss encoders
in our experiments. Both models are pre-trained on released Oxford dictionary data for around 10 epochs. 
We evaluate the model every epoch by the gloss matching accuracy on the randomly divided evaluation set. 
In the pre-training process, we set the GR loss weight as $\lambda=2.0$. 
We use cosine similarity between gloss embedding and target word contextual embedding. As the setting in SimCSE \cite{gao2021simcse} training process, we also use the temperature $\tau=0.05$ in softmax. Taking the MT GR model as an example, 
the softmax of the gloss matching is $\text{softmax}(\text{cosine}(\bm{h}_s, \bm{e}_t) / \tau)$.

We conduct the pre-training on 8 Tesla V100 GPUs. For each GPU, the batch size (related to in-batch negative sampling) is set as 48 for BERT and 36 for RoBERTa model. The learning rate is set $2\times 10^{-5}$ with warm-up setting in the first $10\%$ training steps. The AdamW optimizer is used in the training with default hyper-parameters.

\begin{table*}[]
	\resizebox{\textwidth}{!}{%
		\begin{tabular}{lp{15cm}}
			\\ \hline
			target word & tell \\
			sentence & He held Obi-Wan loosely , gently stroking his back He knew now that it did n't matter what Sampris said , or what Yoda \textbf{told} him . \\
			labels & said to (4), inform (2) \\
			RoBERTa & teach, say, give, call, have \\
			SC GR-RoBERTa & teach, say, warn, instruct, promise \\
			+ post-process &  inform, teach, warn, say, instruct \\
			\hline
			
			target word & think \\
			sentence & Shafer \textbf{thinks} we’re going to cry , “he doesn’t get it!” in reply to his piece” “it” being the amazing world of the Web and new media . \\
			labels & believe (3), feel (1), suspect (1), reckon (1), assume (1) \\
			RoBERTa & say, know, hop, believe, worry \\
			SC GR-RoBERTa &believe, say, hop, expect, suspect  \\
			+ post-process & believe, say, hop, expect, know \\
			\hline 
			
			target word & thus \\
			sentence & The kind of control he exercises is \textbf{thus} likely to be limited to " passive " control such as inspection of produced goods and testing to insure that quality standards are being met . \\
			labels & therefore (5), accordingly (1), consequently (1) \\
			RoBERTa & typically, therefore, then, so, similarly \\
			SC GR-RoBERTa & therefore, consequently, so, accordingly, hence \\
			+ post-process & therefore, consequently, hence, thereby, so \\
			\hline
			
			target word & clean \\
			sentence & Dog and horse owners should be encouraged to \textbf{clean} up after their animals . \\
			labels & scrape (1), clear (2), tidy (2) \\
			RoBERTa & wash, pick, wake, keep, clear \\
			SC GR-RoBERTa & groom, walk, look, care, do \\
			+ post-process & tidy, wash, groom, care, walk \\
			\hline
			
			target word & late \\
			sentence & We were \textbf{late} doing this since I refused to use someone else 's " shopping cart " system that I did n't write and could n't trust . \\
			labels & delayed (3), tardy (2), behind schedule (1), behind time (1), behind (1) \\
			RoBERTa & also, early, just, still, already \\
			SC GR-RoBERTa & early, slow, not, long, behindo \\
			+ post-process & early, slow, prematurely, long, not \\
			\hline
			
			target word & new \\
			sentence &  The lecture itself went well , but a \textbf{new} problem arose .\\
			labels & different (1), extra (1), additional (1), fresh (4)  \\
			RoBERTa & different, big, small, fresh, great \\
			SC GR-RoBERTa & fresh, big, previous, further, different \\
			+ post-process & fresh, renewed, different, previous, recent \\
			\hline

		\end{tabular}%
	}
 \caption{Examples from LS07 benchmark to show the task and model outputs. The number follows each label is the frequency count indicating the number of annotators that provided this substitute. For each model, we report the top 5 candidates in the first 50 predictions in lemmatized form.} 
\label{tab:ls_example}
\end{table*}


\section{Lexical Substitution Details}
\label{appx:ls_example}
As \citet{arefyev-etal-2020-always} reported, 
the process on the format of word candidates influences the metrics. 
We thus (almost) follow their code\footnote{\url{https://github.com/Samsung/LexSubGen}}
and fix the word process in all experiments.
In our experiments, the word process includes lemmatization (\textit{went->go}), 
filtering the candidates having the same lemmatization output with the original word 
and removing duplicate lemmatization of candidates. 
Additionally we filter out the candidates according to the PoS information. 
For example, the word \textit{good} can be used as \textit{noun} or \textit{adj}, 
but it would be unreasonable to serve as  \textit{verb}. We then check the possible PoSs 
for each candidate and filter those words with unmatched PoS with the target word.

In the post-process, the hyper-parameters in (+emb) and validation are tuned in LS07 data. 
Follow the implementation of \citet{arefyev-etal-2020-always}, 
we use cosine similarity and the temperature for similarity is set $1/15$ in all our experiments. 
For the validation process, we follow the idea of \citet{zhou-etal-2019-bert},
but use BERT-base uncased model for validation. 
Following their work, we pick the first 50 candidates to re-rank 
(it has little influence when the number is above 20 in our experiments). 
The values in propose and validate scores are in different scales, 
as one is from logits and the other is from cosine similarity. 
We then adjust the weight of propose score to let its standard deviation be in the same level with the cosine similarity. 
We set the weight as 0.009 for RoBERTa and 0.004 for BERT. 

Table~\ref{tab:ls_example} gives examples of LS task and compares our model outputs with the baseline.

\section{Sentence Similarity}
\label{appx:sent_similarity}
We extend the contextual token similarity measurement into sentence similarity. 
As stated in \cite{li2020sentence}, the dot product similarity between sentence representations
$\bm{h}_c^\top \bm{h}_{c'}$ is difficult to derived theoretically, since it is not explicitly involved
in the BERT pre-training process. Therefore, inspired by token-level lexical substitution task using contextual
probability distribution, 
we consider the probability distribution of a sentence $\textbf{s}_1$ given another sentence $\textbf{s}_2$,
i.e. $p(\textbf{s}_1|\textbf{s}_2)$.

\begin{proposition}\label{prop:chain}
Let $w_1,\cdots, w_n$ be $n$ tokens sampled from a sentence $\textbf{s}$, and $\textbf{c}_i$ be the rest of 
tokens in $\textbf{s}$ except for $w_i$. Let $x_1,\cdots, x_n$ denote
the tokens that can replace $w_1,\cdots, w_n$ in $\textbf{s}$, respectively. 
The joint probability
distribution of $x_1,\cdots, x_n$ given $\textbf{s}$ is formulated as
\begin{equation}\label{eqn:prop}
\log p(x_1,\dots,x_n|\textbf{s})=\\
\sum_{i=1}^n P_i,
\end{equation}
where 
\begin{equation}\label{eq:pi}
P_i=\log p(x_i|\textbf{c}_i,x_{<i})+\text{PMI}(x_i;w_i|\textbf{c}_i, x_{<i}),
\end{equation}
and $x_{<i}$ denotes $x_1, \cdots, x_{i-1}$.
\end{proposition}
\paragraph{Proof}
We use the mathematical induction to proof the proposition. 

When $n=1$, $\log p(x_1|\textbf{s})=P_1$ is equivalent as Eqn. (\ref{pmi_decomp}). 

When $n>1$, we make an assumption that Eqn. (\ref{eqn:prop}) holds true for $n=k-1$, i.e.
$\log p(x_{<k}|\textbf{s})=\sum_{i=1}^{k-1} P_i$.
Then, 
\begin{align}
&\log p(x_{<k}, x_k|\textbf{s}) \nonumber \\
=&\log p(x_k|\textbf{c}_k,x_{<k})+\log\frac{p(x_k|w_k,\textbf{c}_k,x_{<k})}{p(x_k|\textbf{c}_k, x_{<k})} 
\cdots \nonumber\\
&\quad\quad\quad\quad\quad\quad\quad
 +\log \frac{p(x_k, x_{<k}|w_k,\textbf{c}_k)}{p(x_k|w_k,\textbf{c}_k,x_{<k})}\nonumber \\
=&\log p(x_k|\textbf{c}_k,x_{<k})+\text{PMI}(x_k;w_k|\textbf{c}_k,x_{<k}) \cdots \nonumber \\
&\quad\quad\quad\quad\quad\quad\quad\quad + \log p(x_{<k}|\textbf{s}) \nonumber \\
=& P_k + \sum_{i=1}^{k-1} P_i = \sum_{i=1}^{k} P_i,
\end{align}
which means Eqn. (\ref{eqn:prop}) is also true for $n=k$. $\square$

Proposition \ref{prop:chain} indicates one sentence can be transformed into another sentence through
a series of token substitution operations, and the sentence transforming probability can be decomposed
into the sum of a series of contextual token probabilities and contextual token similarities, i.e.
\begin{equation}\label{eq:sent_prob}
p(\textbf{s}_1|\textbf{s}_2)=\sum_{i=1}^n P_i,
\end{equation}
where $P_i$ is defined in Eqn. (\ref{eq:pi}), and $\textbf{s}_1=[x_1,\cdots, x_n], \textbf{s}_2=[w_1,\cdots,w_n]$.
We ignore the case when $\textbf{s}_1$ and $\textbf{s}_2$ have different lengths, since a simple solution
is to pad the shorter sentence to the length of the longer one.

Eqn. (\ref{eq:sent_prob}) and (\ref{eq:pi}) show that the sentence-level 
tasks also benefits from our gloss regularizer,
since the contextual token similarity modeled by gloss matching task also contributes to sentence representation.

\end{document}